\documentclass{amia}
\usepackage{graphicx}
\usepackage[labelfont=bf]{caption}
\usepackage[superscript,nomove]{cite}
\usepackage{color}
\usepackage{enumitem}
\usepackage{multirow}
\usepackage{textcomp}

\newcommand{\bx}[0]{{\textbf x}}

\begin{document}

\title{A Factored Generalized Additive Model for Clinical Decision Support in the Operating Room}

\author{Zhicheng Cui$^{1}$, Bradley A Fritz, MD$^{2}$, Christopher R King, MD PhD$^{2}$, Michael S Avidan, MBBCh$^{2}$, Yixin Chen, PhD$^{1}$}

\institutes{
    $^1$Department of Computer Science and Engineering, Washington University in St Louis, St Louis, MO; $^2$Department of Anesthesiology, Washington University in St Louis, St Louis, MO\\
}

\maketitle

\noindent{\bf Abstract}

\textit{
Logistic regression (LR) is widely used in clinical prediction because it is simple to deploy and easy to interpret. 
Nevertheless, being a linear model, LR has limited expressive capability and often has unsatisfactory performance. 
Generalized additive models (GAMs) extend the linear model with transformations of input features, though feature interaction is not allowed for all GAM variants. 
In this paper, we propose a factored generalized additive model (F-GAM) to preserve the model interpretability for targeted features while allowing a rich model for interaction with features fixed within the individual.
We evaluate F-GAM on prediction of two targets, postoperative acute kidney injury and acute respiratory failure, from a single-center database.
We find superior model performance of F-GAM in terms of AUPRC and AUROC compared to several other GAM implementations, random forests, support vector machine, and a deep neural network.
We find that the model interpretability is good with results with high face validity.}

\section*{Introduction}
Patients undergoing surgery and anesthesia experience external stresses that place them at risk for numerous complications, including acute kidney injury and acute respiratory failure.  
One of the roles of the anesthesia clinician is to regulate the patient's physiology to minimize these risks.  
Logistic regression-based models for predicting postoperative acute kidney injury \cite{chertow1997preoperative,kheterpal2009development,palomba2007acute} and acute respiratory failure\cite{arozullah2000multifactorial,gupta2011development,johnson2007multivariable} have been developed by multiple groups.  
Linear models offer a high degree of transparency regarding which features drive the output, but they are inherently limited in their flexibility, which limits their predictive accuracy.
Various machine learning (ML) models have been proposed to solve these clinical prediction tasks with greater accuracy.  
Classifiers for acute kidney injury have been described in postoperative \cite{thottakkara2016application} and non-surgical hospitalized patients\cite{kate2016prediction,koyner2018development}.  
Although these ML models outperform logistic regression, they are not frequently used in clinical practice in part due to their lack of interpretability.

An interpretable model must provide predictions that are both accountable and actionable.  
An accountable model provides information about which features are contributing to the output prediction.  
This information can include feature importance and feature interactions.  
An actionable model provides guidance regarding how to modify the input features so that the post-intervention features will lead to the desired output.

Interpreting ML models is an extremely active field \cite{caicedo2019iseeu,sha2017interpretable,guidotti2018survey}.  
Incorporating interpretability constraints directly into the structure of the model and using post-hoc interpretation methods are two of the main directions\cite{montavon2018methods}. 
Most existing work focuses on accountability. 
Che et al \cite{che2016interpretable} transfer the DNN's knowledge to gradient boosting trees (GBT) using knowledge distillation and interpret feature importance through measures of variable importance designed for GBT. 
Another work \cite{rajaraman2018visualization} visualizes the region of interest through class activation maps. Ge et al.\cite{ge2018interpretable} feed features extracted from recurrent neural networks into a logistic regression model for prediction, where importance of the transformed features can be directly read off. 
Neural networks with attention-like mechanisms are also popular to visualize the features which contribute the most to a classifier for a particular case \cite{sha2017interpretable}. 
A smaller number of techniques address the actionability requirement. 
An early work \cite{cui2015optimal} proposed an integer linear programming method to extract actionable knowledge from a random forest. 
Gardner et al.\ proposed a label changing method by searching semantically meaningful changes to an image under its manifold space\cite{gardner2015deep}. 
As far as we know, little work has been done on addressing accountability and actionability at the same time in the clinical area.

An extension to generalized linear models, generalized additive models (GAMs), can address accountability and actionability simultaneously. 
Examples includes LR, density based logistic regression \cite{chen2013density} (DLR), generalized additive neural networks \cite{potts1999generalized} (GANN),  
and deep embedding logistic regression \cite{8588790} (DELR). 
LR assumes a fixed (up to a parameter) monotonic relationship between each feature and the outcome probability, limiting its flexibility and predictive performance.
GAMs loosen these assumptions by inferring a transformation of the inputs with the full flexibility of non-parametric or parametric methods.
For example, DLR transforms each feature through a kernel estimator, and 
our recently proposed DELR performs feature-wise nonlinear transformation using neural networks. 
GANN, which used a single hidden layer, can be treated as a special case of DELR, which used multi-layer DNNs.
Despite the increased flexibility, with suitable constraints we can extract accountability and actionability from GAMs. 
Given an input example, GAMs allow us to calculate the contribution of each feature to that example's predicted value. 
Feature contribution curves can be drawn to provide actionable directions on the optimal change and magnitude of improvement for each numeric feature. 
However, only when all the features are conditionally independent (given the label) can GAMs model the true distribution of data\cite{levy2012probabilistic}. 
Feature interactions are not allowed in GAMs, restricting their performance in dealing with complex datasets. In addition, GAMs have the undesirable property of treating static and time-varying features equally.  
For example, demographic characteristics such as age, gender, and height are not possible to change. 
On the other hand, it is possible to deliver interventions that modify a patient's vital signs during surgery.

To address these problems, we propose a variation of GAMs that splits features into time-varying (or targeted) features and static features.
F-GAM fits a context-based scaling for each time-varying factor based on the static factors, substantially increasing its flexibility compared to models which require the effect of a feature to be the same for all examples, but retains the ability to derive personalized feature-effect curves.
F-GAM retains the full flexibility of a DNN for the effect of static features and DNN-based flexibility for the transformation of time-varying factors.
We implement F-GAM as an end-to-end trained model with minimal hyper-parameters.
In extreme cases where there are no static features available, F-GAM reduces to DELR. 
If there are no time-varying features, F-GAM becomes a DNN. 
We empirically validate the accuracy performance of F-GAM with existing ML models, including other GAMs
and demonstrate the interpretability of F-GAM through a case study on predicting acute kidney injury.

\section*{Background and Notation}

\emph{Notation}

Operating room data contains both preoperative data such as demographic information and intraoperative data such as vital signs and medications administered. 
Given a patient $i$, pre-op data $\mathbf{x}^S_i \in \mathcal{R}^{D_1}$ collected before the surgery are treated as static feature vectors while intra-op data represented as $\mathbf{x}_i^{TV} \in \mathcal{R}^{D_2}$ can be modified in real time. 
Together, we use $\mathbf{x}_i = [\mathbf{x}^S_i, \mathbf{x}_i^{TV}] \in \mathcal{R}^{D}$ to denote input features and $y_i \in \{0,1\}$ to represent the binary outcomes. 
 
Our examples are binary classification, but the extension to multi-class classification 
is straightforward with a final softmax transformation and appropriate loss function.

\emph{Generalized Additive Models} 

A generalized additive model (GAM) is an ensemble of $D$ univariate functions, where $D$ is the number of features.
We use $x_{j}$ and $y$ to denote the $j$th dimension of input $\mathbf{x}$ and class label, respectively. 
The output of each univariate function, denoted as $f_t(x_{j})$ is a real number. We can write the GAM structure as
\begin{equation}
	g(E(y))= \beta_0 + f_1(x_{1}) + f_2(x_{2}) + \cdots + f_D(x_{D}),
	\label{eq.gam}
\end{equation}
where the function $g$ is the link function, bounding the range of right hand side value of Eq.\ref{eq.gam}, and $E(y)$ is the expected value of the label conditional on $\bx$. 
Constraints on $f_k$ such as smoothness or degrees of freedom regularize the estimation problem to decrease out-of-sample loss.
With a little abuse of notations, we use $F(\bx)$ to denote $E(y|\bx)$ throughout this paper for ease of presentation. 
By inversing the link function, the GAM has the form,
\begin{equation}
	F(\bx)= g^{-1} [\beta_0 + f_1(x_{1}) + f_2(x_{2}) + \cdots + f_D(x_{D}) ],
	\label{eq.gam_output}
\end{equation}
where the model output is controlled by the sum of each univariate function. 
GAM assumes all the features of input $\bx_i$ are making contributions independently. 
Interpreting a GAM is straightforward as the marginal impact of a specific feature does not rely on the rest of features; 
we are able to know the importance of a feature by plotting its corresponding univariate function or calculating its variance over the sample. 
Actionable changes can be made based the shape of each $f_k(\bx_k)$.

Logistic regression is a special case of GAM by choosing logit function $g(x) = \ln\frac{x}{1-x}$ as the link function and setting $f_k(x_k)$ to be $w_k x_k$ yielding
\begin{equation}
	F(\bx)= \sigma(w_0 + w_1 x_{1} + w_2 x_{2} + \cdots + w_D x_{D}),
	\label{eq.lr}
\end{equation}
where the sigmoid function $\sigma(x) = \frac{1}{1+\exp(-x)}$ is the inverse form of the logit function.
LR assumes a monotonic relationship between the final output $F(\bx)$ and input features due to the linear function $f_k$. 
However, this condition doesn't hold in many cases, such as the relationship between ICU transfer rate and age\cite{chen2013density}. 

\section*{Methods - Model Algorithm}
\begin{figure}[t]
\centering
\includegraphics[scale=0.6]{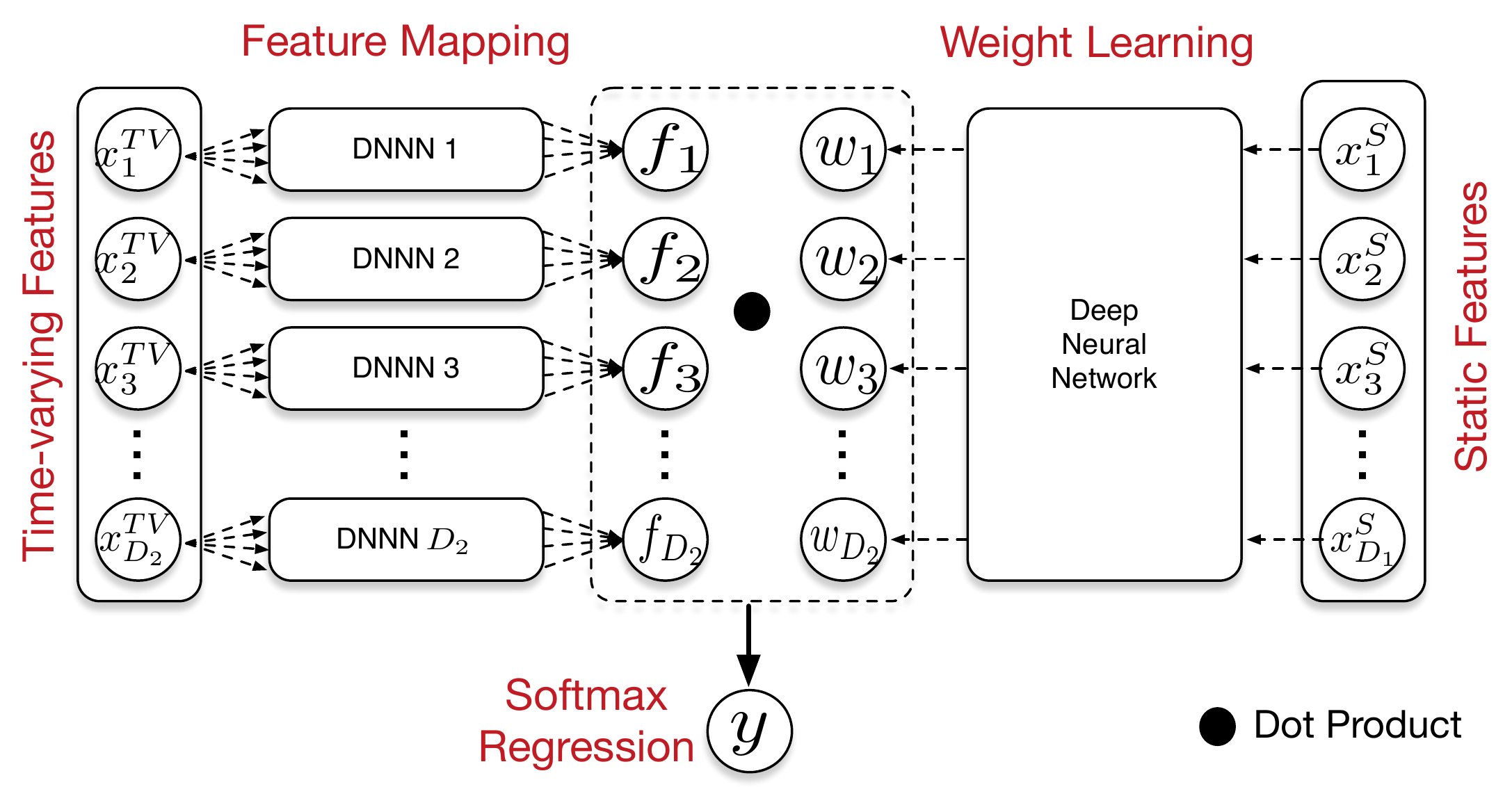}
\caption{Overall architecture of F-GAM. Each circle denotes a scalar. Upper left part is feature mapping module. Every time-varying feature is fed to its own deep and narrow neural network (DNNN) separately. Weight learning module, which is shown in upper right part takes static features as input and calculates feature weights. Note that bias learning module is not plotted in this figure for simplicity.}
\label{fig.arch}
\end{figure}

In this section, we propose a factored generalized additive model (F-GAM) framework in which interactions between time-varying features and static features are allowed.  
The overall model has the form
\begin{equation}
	F(\bx)= \sigma \left [ \sum_{t=1}^{D_2} w_t(\mathbf{x}^S) f_t(x_t^{TV}) + w_0(\mathbf{x}^S) \right ]
	\label{eq.cdelr}
\end{equation}
 In F-GAM, $w_t$ is no longer a constant weight parameter, but the output of a DNN that accepts the static feature vector as input and estimates the weight of $t$th time-varying feature for each case. 
 The feature-wise nonlinear transformation functions $f_t$, $t=1,2,...D_2$ are jointly estimated. 
 $w_0$ is a bias / intercept term that also depends only on the static features.
 In our operative examples, $w_0$ represents the estimate of risk before any intra-operative data becomes available as long as the input features have been appropriately centered.

F-GAM can be decomposed into four different modules: time-varying feature mapping module, feature weights learning module, bias term learning module and logistic/softmax regression module. 
We display the F-GAM architecture in Figure \ref{fig.arch}. 

 \emph{Time-varying feature mapping module} 
 
 In traditional GAMs, the ability of the univariate function $f_k$ to approximate the unknown transformation plays a crucial role in model performance. 
 We choose to use deep and narrow neural networks (DNNN)\cite{8588790} for the nonlinear feature embedding. 
 Being a universal approximator, a DNNN is able learn complex patterns automatically.  
 The general tools for regularizing neural networks are immediately available to control overfitting without the difficult-to-understand smoothness or degree-of-freedom constraints of other GAM transformations.
Each time-varying feature is fed into a DNNN with distinct parameters; 
however, several hyperparameters (depth, width, dropout, training stopping time) are shared across $t$ to avoid having to search over a large hyperparameter space.
The shared architectural parameters also tend to prevent over-fitting of just a few features (data not shown).
A learnable look-up table (categorical embedding) is attached before a DNNN for categorical features. 
In our examples, all time-varying features are quantitative or ordinal rather than categorical.

 \emph{Feature weights learning module}
 
 Rather than applying fixed weights for the input features, we use nonlinear functions $w_t$ to adjust the feature weight dynamically. 
 The nonlinear function should have the following two properties. 
 First, the nonlinear function should not increase the number of parameters dramatically. 
 Second, the nonlinear function should be able to handle both numerical features and categorical ones. 
 Thus, we choose to use deep neural networks as the nonlinear functions. 
 Rather than assigning each $w_t$ a standalone DNN as we did for $f_t$, all the weight-learning functions are estimated with a common DNN except the last layer. 
 With this multi-task setup, we are able to exploit the shared structure of the data to reduce the effective number of parameters. 
 Joint predictions of $w_t$ also allow the module to dynamically choose between potentially correlated $x^M_t$ to emphasize, meaning that $w_t$ represents both the relevance and precision of $f_t$ in the given context. For the second property, we use categorical embedding. 
 When there are no static features, $w_t$ is a constant per time-varying feature and F-GAM reduces to DELR. That is to say, DELR is a special case of our model.

  \emph{Bias term learning module} 
  
  In order to increase expressiveness of the final model, we add a bias term based on the static features. 
  Again, we use a DNN to model the bias term. 
  This DNN is appended to the penultimate layer of the feature weights module to reduce redundancy.
  When there are no time-varying features, only the bias term controls the final output. 
  In this case, F-GAM simplifies to a deep neural network.
  
  \emph{Logistic/softmax regression module} 
  
  With all the transformed features and weights ready, we apply the dot product operation to the time-varying feature mapping and the learned weights. 
  After adding the bias term $w_0$, a sigmoid function $\sigma$ is used to model the positive rate given input data. 
 
  Our F-GAM is trained end-to-end by minimizing the cross entropy loss between true label distribution and prediction distribution. 
  We also apply weight decay and an early stopping strategy to avoid over-fitting. The code is available at https://github.com/nostringattached/FGAM.

\section*{Methods - Experiments}
\emph{Data Sources}

Models were trained and validated using a dataset obtained from a single academic medical center (Barnes Jewish Hospital, St. Louis, Missouri).  All adult patients who received surgery with anesthesia between June 2012 and August 2016 were eligible for inclusion.  Due to the limited incidence of acute respiratory failure among patients who were not admitted to the intensive care unit (ICU) after surgery, prediction of this complication was limited to patients admitted to the ICU after surgery.

Acute kidney injury and acute respiratory failure were the two complications that were used as targets in the experimental models.  Per Kidney Disease: Improving Global Outcomes (KDIGO) criteria, acute kidney injury was defined as an increase in the serum creatinine value by $>$0.3 mg/dL or $>$50\% within 48 hours, compared to the preoperative value\cite{kellum2013diagnosis}.  Acute kidney injury was undefined if the patient was receiving dialysis before surgery.  The preoperative creatinine was the most recent value available before surgery, but no more than 30 days before surgery.  Acute respiratory failure was defined as mechanical ventilation for $>$48 hours after surgery or reintubation within 48 hours.  Acute respiratory failure was undefined if the patient was receiving mechanical ventilation before surgery, if the patient had a second surgery within 48 hours, or if the patient died within 48 hours.

Baseline demographic characteristics, comorbid health conditions, and preoperative laboratory values were retrieved from the electronic medical record.  The total doses of commonly used medications (including intravenous fluids, blood pressure-raising and -lowering agents, sedatives, pain medications, and nephrotoxic antibiotics) were also retrieved. The full list of features included in the analysis is shown in Table \ref{tab.exp}.
\\
\begin{table}[h]
\centering
\caption{Features included in the model.}
\begin{tabular}{|c|c|}
\hline
\begin{tabular}[c]{@{}c@{}}Demographic\\ Characteristics\end{tabular}    & \begin{tabular}[c]{@{}c@{}}Age, Height, Weight, Ideal body weight, Body mass index, Sex, Race, Charlson\\ Comorbidity Index, Functional capacity, American Society of Anesthesiologists\\ physical status, Surgery type\end{tabular}                                                                                                                                                                                                                                                                                                                                                                                             \\ \hline
\begin{tabular}[c]{@{}c@{}}Comorbid\\ Conditions\end{tabular}            & \begin{tabular}[c]{@{}c@{}}Hypertension, Coronary artery disease, Prior myocardial infarction,\\ Congestive heart failure, Diastolic function, Left ventricular ejection fraction, Aortic\\ stenosis, Atrial fibrillation, Pacemaker, Prior stroke, Peripheral artery disease, Deep\\ venous thrombosis, Pulmonary embolism, Diabetes mellitus, Outpatient insulin use,\\ Chronic kidney disease, Ongoing dialysis, Pulmonary hypertension, Chronic obstructive\\ pulmonary disease, Asthma, Obstructive sleep apnea, Cirrhosis, Cancer,\\ Gastro-esophageal reflux, Anemia, Coombs positive, Dementia, Ever-smoker\end{tabular} \\ \hline
Preop Vital Signs                                                        & Systolic blood pressure, Diastolic blood pressure, Pulse oximeter, Heart rate                                                                                                                                                                                                                                                                                                                                                                                                                                                                                                                                                    \\ \hline
Preop Labs                                                               & \begin{tabular}[c]{@{}c@{}}Albumin, Alanine phosphatase, Creatinine, Glucose, Hematocrit, Partial\\ thromboplastin time, Potassium, Sodium, Urea Nitrogen, White blood cells\end{tabular}                                                                                                                                                                                                                                                                                                                                                                                                                                        \\ \hline
\begin{tabular}[c]{@{}c@{}}Intraoperative\\ Time Series\end{tabular}     & \begin{tabular}[c]{@{}c@{}}Mean arterial pressure, Systolic blood pressure, Diastolic blood pressure, Heart rate,\\ Pulse oximeter, Temperature, Respiratory rate, Tidal volume, Peak inspiratory pressure,\\ Positive end-expiratory pressure, Fraction inspired oxygen, End-tidal carbon dioxide,\\ End-tidal anesthetic concentration\end{tabular}                                                                                                                                                                                                                                                                            \\ \hline
\begin{tabular}[c]{@{}c@{}}Intraoperative\\ Meds and Fluids\end{tabular} & \begin{tabular}[c]{@{}c@{}}Albumin, Amiodarone, Crystalloid (lactated ringers + normal saline),\\ Dobutamine, Ephedrine, Epinephrine, Fentanyl, Furosemide, Gentamicin,\\ Hydromorphone, Midazolam, Nicardipine, Norepinephrine, Packed red blood cells,\\ Phenylephrine, Propofol, Remifentanil, Vancomcyin, Vasopressin, Other blood products\end{tabular}                                                                                                                                                                                                                                                                     \\ \hline
\end{tabular}
\label{tab.exp}
\end{table}

For intraoperative time series features, summary measures were derived.  For each feature, the mean, standard deviation, maximum, and minimum over the entire surgery were calculated.  The maximum pulse oximeter reading was omitted due to ceiling effects, while minimum peak inspiratory pressure and minimum tidal volume were omitted due to expected lack of clinical significance.  In addition, the fraction of surgery with extreme values of certain parameters were also calculated, using multiple cutoff values.  These included duration of low mean arterial pressure ($<$55, $<$60, or $<$65 mmHg), high heart rate ($>$100, $>$110, or $>$120 beats per min), low heart rate ($<$60, $<$55, or $<$50 beats per min), low temperature ($<$36 or $<$35.5 \textdegree{}C), low pulse oximeter ($<$90 or $<$85\%), high exhaled carbon dioxide ($>$50 mmHg), low exhaled carbon dioxide ($<$30 mmHg), high peak inspiratory pressure ($>$30 mmHg), and high tidal volume ($>$10 mL per kg).  Lung compliance was also calculated as final tidal volume divided by final peak inspiratory pressure.

\emph{Experimental Technique}

For each of the two target outcomes, F-GAM was compared to four baseline models (decision tree [DT], random forest [RF], support vector machine [SVM], and deep neural network [DNN]) and to three GAMs (logistic regression [LR], gradient boosting decision stumps \cite{hastie01statisticallearning}[GBDS] and deep embedding logistic regression [DELR]).  Note that density based logistic regression (DLR) was not included as it did not finish training in 24 hours. Each model was trained using a 70\%  random sample of the dataset. 10\% of the dataset was selected as a validation set for hyper-parameter tuning and performance was tested on the remaining 20\% of the dataset.  Model performance was quantified using area under the receiver operating characteristic curve (AUROC) and area under the precision-recall curve (AUPRC).  We calculate two-sided 95\% confidence intervals for each measure using the statistical analysis method given by Hanley and McNeil \cite{hanley1982meaning}.

\section*{Results}

\begin{table}[h]
\centering
\caption{AUROC score, AUPRC score and their corresponding 95\% confidence interval (CI) of different methods. DT = decision tree. RF = random forest. SVM = support vector machine, DNN = deep neural network. LR = logistic regression, GBDS = gradient boosting decision stumps, DELR = deep embedding logistic regression, F-GAM = factored generalized additive model.}
\begin{tabular}{|c|c|c|c|c|c|}
\hline
\multicolumn{2}{|c|}{\multirow{3}{*}{Model}} & \multicolumn{2}{c|}{Acute Kidney Injury}                                                                                                                        & \multicolumn{2}{c|}{Acute Respiratory Failure}                                                                                                                  \\ \cline{3-6} 
\multicolumn{2}{|c|}{}                       & \begin{tabular}[c]{@{}c@{}}AUROC\\ 95\% CI\end{tabular}                        & \begin{tabular}[c]{@{}c@{}}AUPRC\\ 95\% CI\end{tabular}                        & \begin{tabular}[c]{@{}c@{}}AUROC\\ 95\% CI\end{tabular}                          & \begin{tabular}[c]{@{}c@{}}AUPRC\\ 95\% CI\end{tabular}                        \\ \hline
\multirow{7}{*}{Baselines}         
                                   & DT      & \begin{tabular}[c]{@{}c@{}}0.580\\ {[}0.563, 0.597{]}\end{tabular}          & \begin{tabular}[c]{@{}c@{}}0.137\\ {[}0.130, 0.145{]}\end{tabular}          & \begin{tabular}[c]{@{}c@{}}0.535\\ {[}0.474, 0.595{]}\end{tabular}          & \begin{tabular}[c]{@{}c@{}}0.043\\ {[}0.033, 0.053{]}\end{tabular}          \\ \cline{2-6} 
                                   & RF      & \begin{tabular}[c]{@{}c@{}}0.820\\ {[}0.806, 0.835{]}\end{tabular}          & \begin{tabular}[c]{@{}c@{}}0.253\\ {[}0.243, 0.266{]}\end{tabular}          & \begin{tabular}[c]{@{}c@{}}0.718\\ {[}0.658, 0.777{]}\end{tabular}          & \begin{tabular}[c]{@{}c@{}}0.085\\ {[}0.068, 0.102{]}\end{tabular}          \\ \cline{2-6} 
                                   & SVM     & \begin{tabular}[c]{@{}c@{}}0.794\\ {[}0.779, 0.809{]}\end{tabular}          & \begin{tabular}[c]{@{}c@{}}0.215\\ {[}0.205, 0.226{]}\end{tabular}          & \begin{tabular}[c]{@{}c@{}}0.698\\ {[}0.638, 0.758{]}\end{tabular}          & \begin{tabular}[c]{@{}c@{}}0.094\\ {[}0.076, 0.113{]}\end{tabular}          \\ \cline{2-6} 
                                   & DNN     & \begin{tabular}[c]{@{}c@{}}0.787\\ {[}0.772, 0.802{]}\end{tabular}          & \begin{tabular}[c]{@{}c@{}}0.216\\ {[}0.206, 0.227{]}\end{tabular}          & \begin{tabular}[c]{@{}c@{}}0.698\\ {[}0.638, 0,758{]}\end{tabular}          & \begin{tabular}[c]{@{}c@{}}0.084\\ {[}0.072, 0.109{]}\end{tabular}          \\ \hline
\multirow{5}{*}{GAMs}              & LR      & \begin{tabular}[c]{@{}c@{}}0.794\\ {[}0.783, 0.813{]}\end{tabular}          & \begin{tabular}[c]{@{}c@{}}0.221\\ {[}0.212, 0.233{]}\end{tabular}          & \begin{tabular}[c]{@{}c@{}}0.650\\ {[}0.052, 0.712{]}\end{tabular}          & \begin{tabular}[c]{@{}c@{}}0.073\\ {[}0.058, 0.088{]}\end{tabular}          \\ \cline{2-6} 

& GBDS    & \begin{tabular}[c]{@{}c@{}}0.803\\ {[}0.788, 0.818{]}\end{tabular}          & \begin{tabular}[c]{@{}c@{}}0.253\\ {[}0.242, 0.265{]}\end{tabular}          & \begin{tabular}[c]{@{}c@{}}0.713\\ {[}0.654, 0.773{]}\end{tabular}          & \begin{tabular}[c]{@{}c@{}}0.084\\ {[}0.070, 0.105{]}\end{tabular}          \\ \cline{2-6} 
                                   & DELR    & \begin{tabular}[c]{@{}c@{}}0.800\\ {[}0.786, 0.815{]}\end{tabular}          & \begin{tabular}[c]{@{}c@{}}0.235\\ {[}0.225, 0.247{]}\end{tabular}          & \begin{tabular}[c]{@{}c@{}}0.708\\ {[}0.648, 0.768{]}\end{tabular}          & \begin{tabular}[c]{@{}c@{}}0.083\\ {[}0.066, 0.099{]}\end{tabular}          \\ \hline
                                   
\multicolumn{1}{|l|}{Our Method}   & F-GAM   & \textbf{\begin{tabular}[c]{@{}c@{}}0.824\\ {[}0.813, 0.842{]}\end{tabular}} & \textbf{\begin{tabular}[c]{@{}c@{}}0.264\\ {[}0.258, 0.282{]}\end{tabular}} & \textbf{\begin{tabular}[c]{@{}c@{}}0.718\\ {[}0.659, 0.777{]}\end{tabular}} & \textbf{\begin{tabular}[c]{@{}c@{}}0.106\\ {[}0.091, 0.134{]}\end{tabular}} \\ \hline
\end{tabular}

\label{tab.result}
\end{table}
The dataset included 111,890 patients.  Of these patients, 5,018 were excluded from the acute kidney injury model because they were receiving dialysis before surgery or because no postoperative creatinine value was available.  Of the remaining 106,872 patients, 6,472 (6.1\%) experienced acute kidney injury.  Of the original 111,890 patients, 89,688 were excluded from the acute respiratory failure model because they were not admitted to the intensive care unit, while 6,578 were excluded due to preoperative mechanical ventilation or one of the other exclusion criteria.  Of the remaining 15,624 patients, 489 (3.1\%) experienced acute respiratory failure.

\begin{figure}[h]
\centering
\includegraphics[scale=0.4]{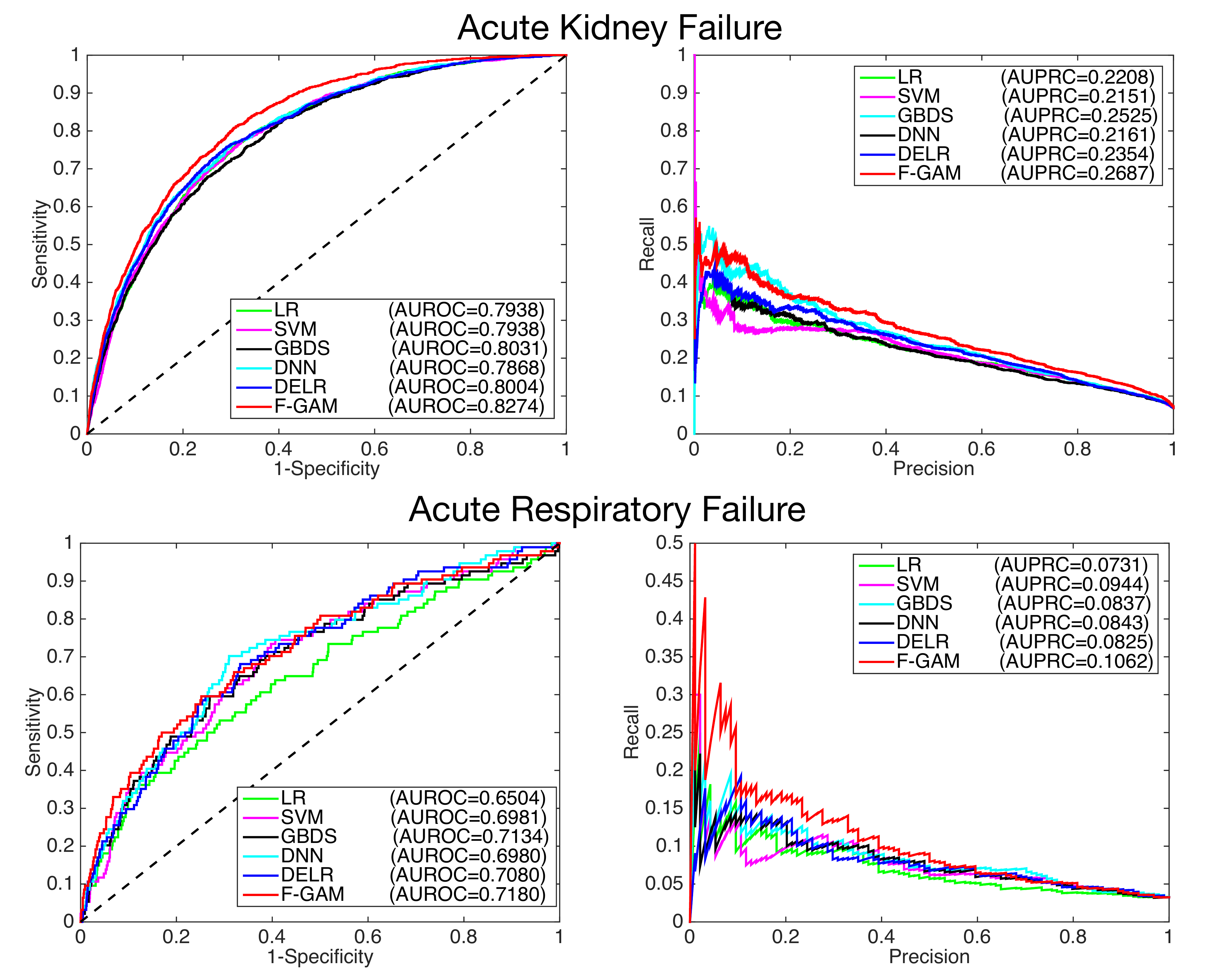}
\caption{ROC curve and precision recall curve (PRC) of different models predicting acute kidney injury and acute respiratory failure. LR = logistic regression, SVM = support vector machine, GBDS = gradient boosting decision stumps, DNN = deep neural network, DELR = deep embedding logistic regression, F-GAM = factored generalized additive model.}
\label{fig.roc_apr}
\end{figure}

 Performance of the models is shown in Table \ref{tab.result}, while the receiver-operating characteristic and precision-recall curves are shown in Figure \ref{fig.roc_apr}. For both outcomes, F-GAM provided the highest AUROC and the highest AUPRC.  DT and RF are excluded from Figure \ref{fig.roc_apr} for readability purposes.

Figure 3 demonstrates how the contribution $w_t (\bx^S) f_t (x_t ^{TV})$ to the predicted risk of acute kidney injury changes at different values $x_t^{TV}$ of four representative time-varying features in two randomly selected patients. Each panel assumes that all other time-varying features remain constant. Points that are higher on the vertical axis represent a larger contribution to the predicted probability. 

\begin{figure}[h]
\centering
\includegraphics[scale=0.5]{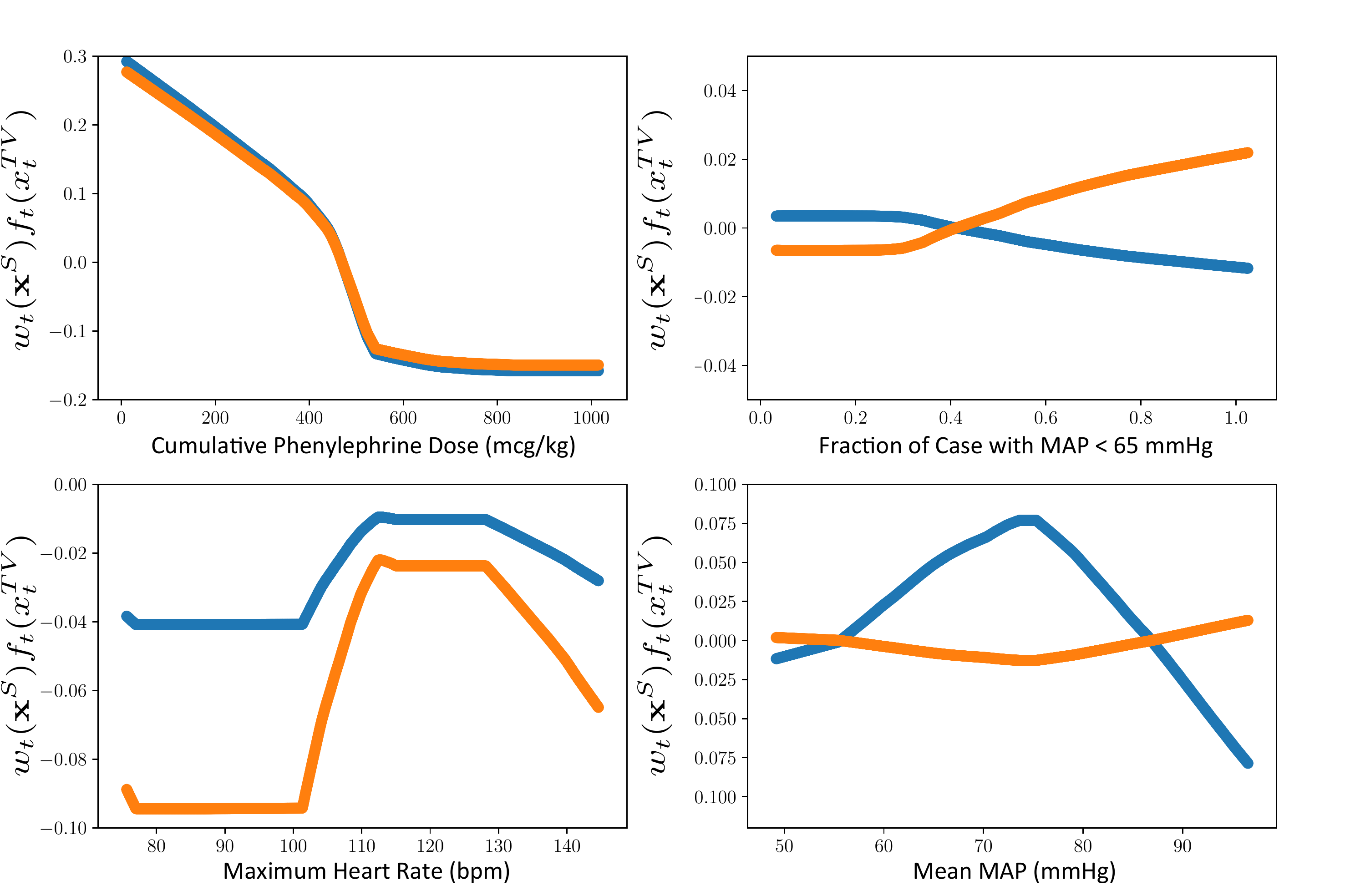}
\caption{Contribution of each feature to the predicted probability of acute kidney injury as a function of feature value. Each panel assumes that all other dynamic features are held constant. The blue curve shows the feature contributions in a 57-year-old healthy female (who ultimately did not have AKI), while the orange curve shows the feature contributions in a 49-year-old female with hypertension, chronic kidney disease, and cirrhosis of the liver (who did have AKI). }
\label{fig.inter}
\end{figure}

\section*{Discussion}
Our experimental results demonstrate that F-GAM outperforms other methods with respect to accuracy on this task while also offering the benefits of accountability and actionability.
All of the models tended to perform better for kidney injury than for respiratory failure, which is likely related to the higher incidence of kidney injury in our dataset and the larger sample size used for this outcome (106,872 versus 15,624).  
The pure deep neural network didn't perform as well in our dataset, likely because it is very easy to overfit despite traditional regularization methods such as learning rate decay and weight decay being applied. 
The random forest model had performance characteristics that were most similar to F-GAM, but F-GAM would be preferable over the random forest because F-GAM offers interpretability, while the random forest does not. 

Figure \ref{fig.inter} demonstrates how predictions generated by F-GAM can be used to guide intraoperative management.  For example, the lower-left panel shows that an increase in maximum heart rate from 100 bpm to 110 bpm appears to be associated with an increased risk for acute kidney injury, because the graph has a steep positive slope in this region. On the other hand, an increase from 80 bpm to 90 bpm is not associated with increased risk, because the graph has a flat slope in this region. Multiplication times the scalar $w_t (\bx^S)$ allows the curve to expand or shrink vertically depending on the baseline characteristics and other health conditions of an individual patient.  Thus an increase in maximum heart rate from 100 bpm to 110 bpm appears to be associated with increased risk both in a healthy patient (blue curve) and in a very ill patient (orange curve), but the increase in risk (i.e., the slope) is much greater for the very ill patient. This observation fits the anesthesia clinician's intuition.

The anesthesia clinician should not assume that the associations reported by this model indicate that elevated heart rate causes acute kidney injury. Nor should the clinician assume that blindly giving a medication that lowers the heart rate will decrease the patient's risk for acute kidney injury. On the contrary, increased heart rate is often a sign of an underlying problem, such as dehydration. The underlying problem, not the fast heart rate, is what increases the risk for acute kidney injury. It is the clinician's job to identify the underlying problem and correct it. 

The upper-left panel provides another example of a scenario where the reported correlation should not be assumed to indicate causation. The graph shows a negative slope as the cumulative dose of phenylephrine (a medication that raises blood pressure) increases from 0 to 100 mcg/kg.  This suggests that administration of low doses of phenylephrine might decrease the risk of acute kidney injury (if all other features remain constant).
This example demonstrates one of the limitations of any factorized model structure; it is unlikely that large doses of phenylephrine decrease the risk of acute kidney injury in and of itself, but it is well supported in the anesthesia literature that untreated low blood pressure increases acute kidney injury risk.
Additionally, zero or very low doses of phenylephrine (a weaker first line drug) may represent immediate escalation to stronger vasopressor such as norephinephrine, which increases risk.

In regions where the curve is relatively flat, observing a different value for that feature will have minimal impact on the predicted probability of the target. This can be seen in the left portion of the upper-right panel. Static features (such as age) may also be important (and our model accounts for these effects through the presence of static features in the weights $w_t (\bx^S)$ and in the bias term), but these features by definition are non-modifiable and change neither when intraoperative problems occur nor when the problems are corrected.

When F-GAM predicts a high probability of an adverse outcome such as acute kidney injury, the time-varying features contributing the most to that prediction are those with the highest current values of $w_t (\bx^S) f_t (x_t ^{TV})$. Curves similar to those shown in Figure 3 can be shown to the clinicians in real time during surgery. When a clinician enters the room to provide assistance with a patient who triggered a high-risk alert, these curves can help the clinician quickly determine what features are important in this particular case.  This saves time that would otherwise be spent reviewing all the vital signs and other data. In an environment with as much real-time data as an operating room, streamlining data review is a major advantage, particularly if the clinician coming to provide assistance is otherwise unfamiliar with the patient. Ideally, the clinician will identify the underlying diagnosis sooner and deliver treatment sooner, if treatment is needed.

\section*{Conclusion}
In this paper, we have described a novel Factored Generalized Additive Model (F-GAM) and demonstrated its use in predicting postoperative acute kidney injury and acute respiratory failure in a historical cohort of patients receiving surgery with anesthesia. F-GAM allows for interactions between static and time-varying input features while retaining the qualities of accountability and actionability. Our model outperformed baseline models and other GAMs in predicting both of the complications tested, and the graphical displays of risk indicated associations that have face validity to an anesthesia clinician.  Next steps include application of this technique to other outcomes and prospective deployment of these models for prediction of complications.

\makeatletter
\renewcommand{\@biblabel}[1]{\hfill #1.}
\makeatother

\bibliographystyle{vancouver}

\bibliography{ref_fgam}
\end{document}